\documentclass[10pt,twocolumn,letterpaper]{article}

\usepackage[usenames,dvipsnames,svgnames,table]{xcolor}
\usepackage{cvpr}
\usepackage{times}
\usepackage{graphicx}
\usepackage{amsmath}
\usepackage{amssymb}
\usepackage{paralist}
\usepackage[numbers,sort&compress]{natbib}
\usepackage{booktabs}
\usepackage{wrapfig}
\usepackage{tikz}
\usepackage{pdfpages}

\graphicspath{{foa-examples/}{generation-examples/}}

\newcommand{\ra}[1]{\renewcommand{\arraystretch}{#1}}
\newcommand{\defoccur}[1]{\textsl{#1}}

\newcommand{\red}{\color{DarkRed}}
\newcommand{\green}{\color{SeaGreen}}

\newcommand{\define}{\:{\scriptstyle\stackrel{\triangle}{=}}\:}
\newcommand{\degr}{^\circ}
\newcommand{\dagr}{^{[3,]}}     
\newcommand{\plus}{^+}

\newcommand{\conjunct}{\wedge}
\newcommand{\disjunct}{\vee}

\DeclareMathOperator*{\normal}{\perp\hspace*{-0.5pt}}

\defaultleftmargin{8pt}{}{}{}

\usepackage[pagebackref=true,breaklinks=true,letterpaper=true,colorlinks,bookmarks=false]{hyperref}
\cvprfinalcopy 

\def\cvprPaperID{1701} 

\ifcvprfinal\fi
\begin{document}

\title{Seeing What You're Told: Sentence-Guided Activity Recognition In Video}

\author{%
  N. Siddharth\\
  Stanford University\\
  {\small\texttt{nsid@stanford.edu}}
  \and
  Andrei Barbu\\
  Massachusetts Institute of Technology\\
  {\small\texttt{andrei@0xab.com}}
  \and
  Jeffrey Mark Siskind\\
  Purdue University\\
  {\small\texttt{qobi@purdue.edu}}}

\includepdf[pages={1}]{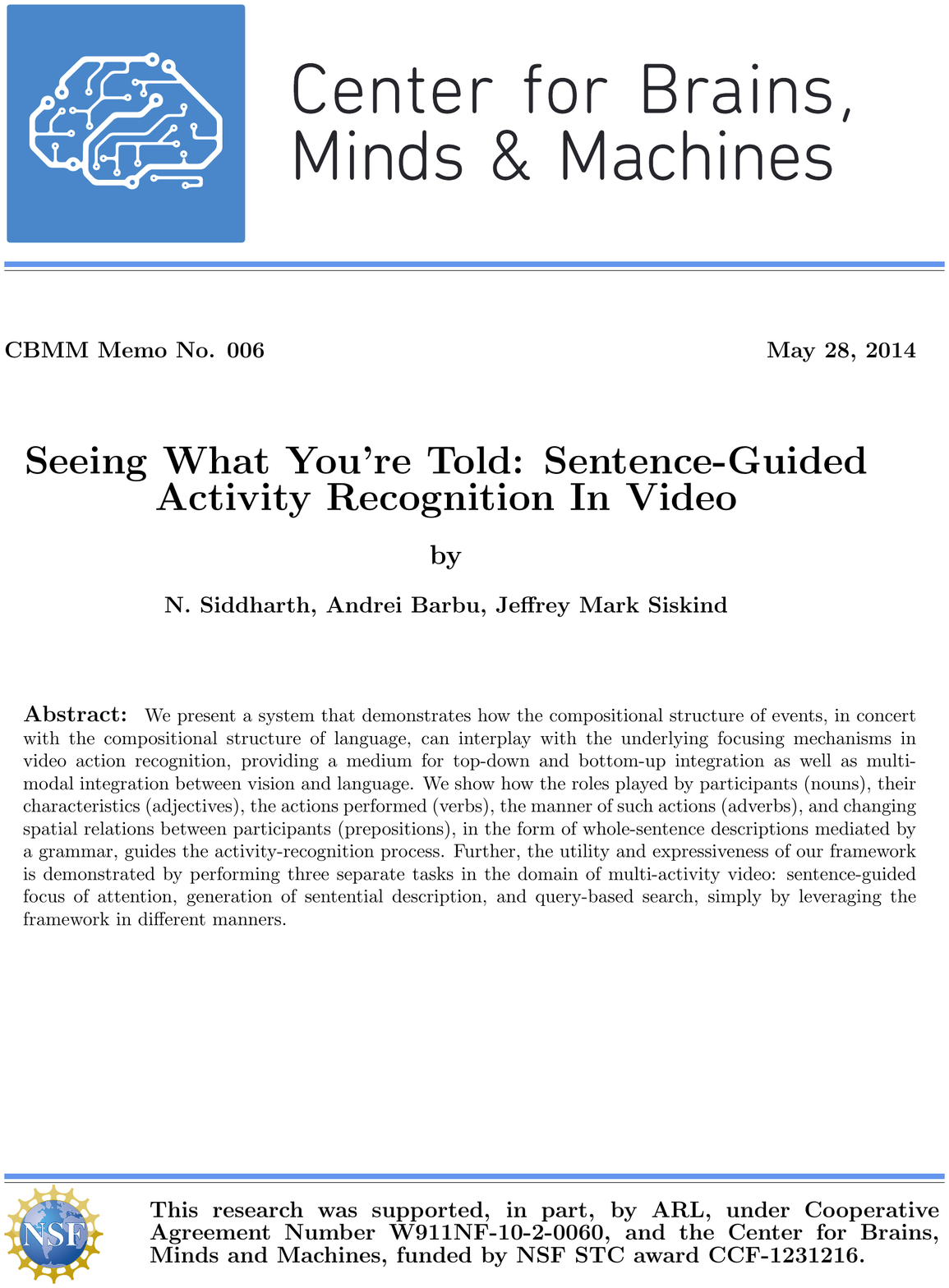}
\setcounter{page}{1}

\maketitle

\begin{abstract}
  \vspace*{-1ex}
  We present a system that demonstrates how the compositional structure of
  events, in concert with the compositional structure of language, can
  interplay with the underlying focusing mechanisms in video action
  recognition, providing a medium for top-down and bottom-up integration as
  well as multi-modal integration between vision and language.
  We show how the roles played by participants (nouns), their characteristics
  (adjectives), the actions performed (verbs), the manner of such actions
  (adverbs), and changing spatial relations between participants (prepositions),
  in the form of whole-sentence descriptions mediated by a grammar, guides
  the activity-recognition process.
  Further, the utility and expressiveness of our framework is demonstrated by
  performing three separate tasks in the domain of multi-activity video:
  sentence-guided focus of attention, generation of sentential description, and
  query-based search, simply by leveraging the framework in different manners.
\end{abstract}

\vspace*{-4ex}
\section{Introduction}
\label{sec:introduction}
\vspace*{-1ex}
The ability to describe the observed world in natural language is a
quintessential component of human intelligence.
A particular feature of this ability is the use of rich sentences, involving
the composition of multiple nouns, adjectives, verbs, adverbs, and
prepositions, to describe not just static objects and scenes, but also events
that unfold over time.
Furthermore, this ability appears to be learned by virtually all children.
The deep semantic information learned is multi-purpose: it supports
comprehension, generation, and inference.
In this work, we investigate the intuition, and the precise means and
mechanisms that will enable us to support such ability in the domain of
activity recognition in multi-activity video.

Suppose we wanted to recognize an occurrence of an event described by the
sentence \emph{The ball bounced}, in a video clip.
Nominally, we would need to detect the \emph{ball} and its position in the
field of view in each frame and determine that the sequence of such detections
satisfied the requirements of \emph{bounce}.
The sequence of such detections and their corresponding positions over time
constitutes a \defoccur{track} for that object.
Here, the semantics of an intransitive verb like \emph{bounce} would be
formulated as a unary predicate over object tracks.
Recognizing occurrences of events described by sentences containing transitive
verbs, like \emph{The person approached the ball}, would require detecting and
tracking two objects, the \emph{person} and the \emph{ball} constrained by a
binary predicate.

In an ideal world, event recognition would proceed in a purely feed-forward
fashion: robust and unambiguous object detection and tracking followed by
application of the semantic predicates on the recovered tracks.
However, the current state-of-the-art in computer vision is far from this
ideal.
Object detection alone is highly unreliable.
The best current average-precision scores on PASCAL VOC hover around 40\%-50\%
\citep{Everingham10}.
As a result, object detectors suffer from both false positives and false
negatives.
One way around this is to use detection-based tracking \citep{Wolf1989}, where
one biases the detector to overgenerate, alleviating the problem of false
negatives, and uses a different mechanism to select among the overgenerated
detections to alleviate the problem of false positives.
One such mechanism selects detections that are temporally coherent, \ie\ the
track motion being consistent with optical flow.
Barbu \etal\ \cite{Barbu2012b} proposed an alternate mechanism that
selected detections for a track that satisfied a unary predicate such as one
would construct for an intransitive verb like \emph{bounce}.
We significantly extend that approach, selecting detections for multiple tracks
that collectively satisfy a complex multi-argument predicate representing the
semantics of an entire sentence.
That predicate is constructed as a conjunction of predicates representing the
semantics of individual words in that sentence.
For example, given the sentence \emph{The person to the left of the chair
  approached the trash can}, we construct a logical form.
\vspace*{-2ex}
\begin{equation*}
  \begin{array}[t]{l}
    \textsc{person}(P)\wedge
    \textsc{toTheLeftOf}(P,Q)\wedge
    \textsc{chair}(Q)\\
    \wedge\ \textsc{approach}(P,R)\wedge
    \textsc{trashCan}(R)
  \end{array}
\vspace*{-0.5ex}
\end{equation*}
Our tracker is able to simultaneously construct three tracks~$P$, $Q$, and~$R$,
selecting out detections for each, in an optimal fashion that simultaneously
optimizes a joint measure of detection score and temporal coherence while also
satisfying the above conjunction of predicates.
We obtain the aforementioned detections by employing a state-of-the-art object
detector \citep{Felzenszwalb2010b}, where we train a model for each object
(\eg\ \emph{person}, \emph{chair}, \etc), which when applied to an image,
produces axis-aligned bounding rectangles with associated scores indicating
strength of detection.

We represent the semantics of lexical items like \emph{person}, \emph{to the
  left of}, \emph{chair}, \emph{approach}, and \emph{trash can} with predicates
over tracks like $\textsc{person}(P)$, $\textsc{toTheLeftOf}(P,Q)$,
$\textsc{chair}(Q)$, $\textsc{approach}(P,R)$, and $\textsc{trashCan}(R)$.
%
%
These predicates are in turn represented as regular expressions
(\ie\ finite-state recognizers or FSMs) over features extracted from the
sequence of detection positions, shapes, and sizes as well as their temporal
derivatives.
For example, the predicate $\textsc{toTheLeftOf}(P,Q)$ might be a single state
FSM where, on a frame-by-frame basis, the centers of the detections for~$P$ are
constrained to have a lower $x$-coordinate than the centers of the detections
for~$Q$.
The actual formulation of the predicates (Table~\ref{tab:predicates}) is more
complex as it must deal with noise and variance in real-world video.
What is central is that the semantics of \emph{all} parts of speech, namely
nouns, adjectives, verbs, adverbs, and prepositions (both those that describe
spatial-relations and those that describe motion), is uniformly represented by
the same mechanism: predicates over tracks formulated as finite-state
recognizers over features extracted from the detections in those tracks.

We refer to this capacity as the \defoccur{Sentence Tracker}, a function
$\mathcal{S}:(\mathbf{B},\mathbf{s},\Lambda)\mapsto(\tau,\mathbf{J})$, that
takes, as input, an overgenerated set~$\mathbf{B}$ of detections along with a
sentence~$\mathbf{s}$ and a lexicon~$\Lambda$ and produces a score~$\tau$
together with a set~$\mathbf{J}$ of tracks that satisfy~$\mathbf{s}$ while
optimizing a linear combination of detection scores and temporal coherence.
This can be used for three distinct purposes as shown in
section~\ref{sec:experiments}:
\begin{compactdesc}
\item[focus of attention] One can apply the sentence tracker to the same
  video clip~$\mathbf{B}$, that depicts multiple simultaneous events taking
  place in the field of view with different participants, with two different
  sentences~$\mathbf{s}_1$ and~$\mathbf{s}_2$.
  In other words, one can compute
  $(\tau_1,\mathbf{J}_1)=\mathcal{S}(\mathbf{B},\mathbf{s}_1,\Lambda)$ and
  $(\tau_2,\mathbf{J}_2)=\mathcal{S}(\mathbf{B},\mathbf{s}_2,\Lambda)$ to yield
  two different sets of tracks~$\mathbf{J}_1$ and~$\mathbf{J}_2$ corresponding
  to the different sets of participants in the different events described
  by~$\mathbf{s}_1$ and~$\mathbf{s}_2$.
\item[generation] One can take a video clip~$\mathbf{B}$ as input and
  systematically search the space of all possible sentences~$\mathbf{s}$ that
  can be generated by a context-free grammar and find that
  sentence~$\mathbf{s}^{*}$ for which
  $(\tau^{*},\mathbf{J}^{*})=\mathcal{S}(\mathbf{B},\mathbf{s}^{*},\Lambda)$
  yields the maximal~$\tau^{*}$.
  This can be used to generate a sentence that describes an input video
  clip~$\mathbf{B}$.
\item[retrieval] One can take a
  collection~$\mathcal{B}=\{\mathbf{B}_1,\ldots,\mathbf{B}_M\}$ of
  video clips (or a single long video chopped into short clips) along with a
  sentential query~$\mathbf{s}$, compute
  $(\tau_i,\mathbf{J}_i)=\mathcal{S}(\mathbf{B}_i,\mathbf{s},\Lambda)$ for
  each~$\mathbf{B}_i$, and find the clip~$\mathbf{B}_i$ with maximal
  score~$\tau_i$.
  This can be used to perform sentence-based video search.
\end{compactdesc}
(Prior work \cite{Yu2013} showed how one can take a training set
$\{(\mathbf{B}_1,\mathbf{s}_1),\ldots,(\mathbf{B}_M,\mathbf{s}_M)\}$ of
video-sentence pairs, where the word meanings~$\Lambda$ are unknown, and
compute the lexicon~$\Lambda^{*}$ which maximizes the sum
$\tau_1+\cdots+\tau_M$ computed from
$(\tau_1,\mathbf{J}_1)=\mathcal{S}(\mathbf{B}_1,\mathbf{s},\Lambda^{*}),\ldots,(\tau_M,\mathbf{J}_M)=\mathcal{S}(\mathbf{B}_M,\mathbf{s},\Lambda^{*})$.)
However, we first present the two central algorithmic contributions of this
work.
In section~\ref{sec:tracker} we present the details of the sentence tracker,
the mechanism for efficiently constraining several parallel detection-based
trackers, one for each participant, with a conjunction of finite-state
recognizers.
In section~\ref{sec:semantics} we present lexical semantics for a small
vocabulary of~17 lexical items (5~nouns, 2~adjectives, 4~verbs, 2~adverbs,
2~spatial-relation prepositions, and 2~motion prepositions) all formulated as
finite-state recognizers over features extracted from detections produced by an
object detector, together with compositional semantics that maps a sentence to
a semantic formula constructed from these finite-state recognizers where
the object tracks are assigned to arguments of these recognizers.

\vspace*{-2ex}
\section{The Sentence Tracker}
\label{sec:tracker}
\vspace*{-1ex}

Barbu \etal\ \cite{Barbu2012b} address the issue of selecting detections
for a track that simultaneously satisfies a temporal-coherence measure and a
single predicate corresponding to an intransitive verb such as \emph{bounce}.
Doing so constitutes the integration of top-down high-level information, in the
form of an event model, with bottom-up low-level information in the form of
object detectors.
We provide a short review of the relevant material in that work to introduce
notation and provide the basis for our exposition of the sentence tracker.
\vspace*{-2ex}
\begin{equation}
  \max_{j^1,\ldots,j^T}
  \sum_{t=1}^Tf(b^t_{j^t})+\sum_{t=2}^Tg(b^{t-1}_{j^{t-1}},b^t_{j^t})
  \label{eq:tracker}
\vspace*{-1ex}
\end{equation}
The first component is a detection-based tracker.
For a given video clip with~$T$ frames, let~$j$ be the index of a detection
and~$b^t_j$ be a particular detection in frame~$t$ with score $f(b^t_j)$.
A sequence $\langle j^1,\ldots,j^T\rangle$ of detection indices, one for each
frame~$t$, denotes a track comprising detections $b^t_{j^t}$.
We seek a track that maximizes a linear combination of aggregate detection
score, summing $f(b^t_{j^t})$ over all frames, and a measure of temporal
coherence, as formulated in Eq.~\ref{eq:tracker}.
The temporal coherence measure aggregates a local measure~$g$ computed between
pairs of adjacent frames, taken to be the negative Euclidean distance between
the center of $b^t_{j^t}$ and the forward-projected center of
$b^{t-1}_{j^{t-1}}$ computed with optical flow.
Eq.~\ref{eq:tracker} can be computed in polynomial time using
dynamic-programming with the Viterbi \cite{Viterbi1971} algorithm.
It does so by forming a lattice, whose rows are indexed by~$j$ and whose
columns are indexed by~$t$, where the node at row~$j$ and column~$t$ is the
detection $b^t_j$.
Finding a track thus reduces to finding a path through this lattice.
\vspace*{-2ex}
\begin{equation}
  \max_{k^1,\ldots,k^T}
  \sum_{t=1}^Th(k^t,b^t_{\hat{\jmath}^t})+\sum_{t=2}^Ta(k^{t-1},k^t)
  \label{eq:map}
\vspace*{-2ex}
\end{equation}
The second component recognizes events with hidden Markov models (HMMs), by
finding a MAP estimate of an event model given a track.
This is computed as shown in Eq.~\ref{eq:map}, where~$k^t$ denotes the
state for frame~$t$, $h(k,b)$ denotes the $\log$ probability of generating a
detection~$b$ conditioned on being in state~$k$, $a(k',k)$ denotes the $\log$
probability of transitioning from state~$k'$ to~$k$, and $\hat{\jmath}^t$
denotes the index of the detection produced by the tracker in frame~$t$.
This can also be computed in polynomial time using the Viterbi algorithm.
Doing so induces a lattice, whose rows are indexed by~$k$ and whose columns are
indexed by~$t$.

The two components, detection-based tracking and event recognition, can be
merged by combining the cost functions from Eq.~\ref{eq:tracker} and
Eq.~\ref{eq:map} to yield a unified cost function
\vspace*{-2ex}
\begin{equation*}
  \max_{\substack{j^1,\ldots,j^T\\k^1,\ldots,k^T}}
  \begin{array}[t]{l}
  \displaystyle \sum_{t=1}^T f(b^t_{j^t})+\sum_{t=2}^T g(b^{t-1}_{j^{t-1}},b^t_{j^t})\\
  \displaystyle {}+\sum_{t=1}^Th(k^t,b^t_{j^t})+\displaystyle\sum_{t=2}^Ta(k^{t-1},k^t)
  \end{array}
\vspace*{-1ex}
\end{equation*}
that computes the joint MAP estimate of the best possible track and the best
possible state sequence.
This is done by replacing the $\hat{\jmath}^t$ in Eq.~\ref{eq:map} with~$j^t$,
allowing the joint maximization over detection and state sequences.
This too can be computed in polynomial time with the Viterbi algorithm, finding
the optimal path through a cross-product lattice where each node represents a
detection paired with an event-model state.
This formulation combines a single tracker lattice with a single event model,
constraining the detection-based tracker to find a track that is not only
temporally coherent but also satisfies the event model.
This can be used to select that \emph{ball} track from a video clip that
contains multiple balls that exhibits the motion characteristics of an
intransitive verb such as \emph{bounce}.

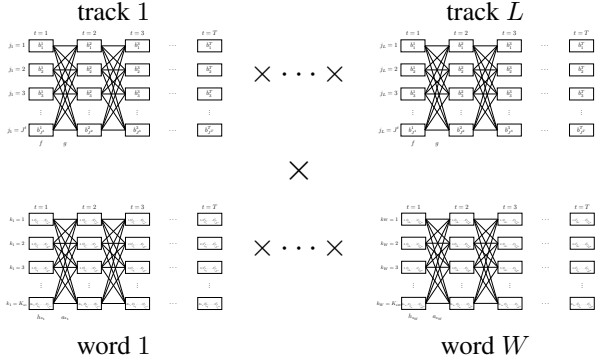
\begin{figure}[t]
  \begin{center}
    \begin{tabular}{@{}c@{\hspace*{10pt}}c@{\hspace*{10pt}}c@{}}
      track $1$&&track $L$\\
      \scalebox{0.2}{
        \begin{tikzpicture}[y=-1cm]

\draw[black] (3.8,4.2) -- (5.4,4.2);
\draw[black] (3.8,2.6) -- (5.4,2.6);
\draw[black] (3.8,5.8) -- (5.4,5.8);
\draw[black] (3.8,8.2) -- (5.4,8.2);
\draw[black] (3.8,2.6) -- (5.4,4.2);
\draw[black] (3.8,2.6) -- (5.4,5.8);
\draw[black] (3.8,2.6) -- (5.4,8.2);
\draw[black] (3.8,4.2) -- (5.4,2.6);
\draw[black] (3.8,4.2) -- (5.4,5.8);
\draw[black] (3.8,4.2) -- (5.4,8.2);
\draw[black] (3.8,5.8) -- (5.4,2.6);
\draw[black] (3.8,5.8) -- (5.4,4.2);
\draw[black] (3.8,5.8) -- (5.4,8.2);
\draw[black] (3.8,8.2) -- (5.4,2.6);
\draw[black] (3.8,8.2) -- (5.4,4.2);
\draw[black] (3.8,8.2) -- (5.4,5.8);
\draw[black] (7,4.2) -- (8.6,4.2);
\draw[black] (7,2.6) -- (8.6,2.6);
\draw[black] (7,5.8) -- (8.6,5.8);
\draw[black] (7,8.2) -- (8.6,8.2);
\draw[black] (7,2.6) -- (8.6,4.2);
\draw[black] (7,2.6) -- (8.6,5.8);
\draw[black] (7,2.6) -- (8.6,8.2);
\draw[black] (7,4.2) -- (8.6,2.6);
\draw[black] (7,4.2) -- (8.6,5.8);
\draw[black] (7,4.2) -- (8.6,8.2);
\draw[black] (7,5.8) -- (8.6,2.6);
\draw[black] (7,5.8) -- (8.6,4.2);
\draw[black] (7,5.8) -- (8.6,8.2);
\draw[black] (7,8.2) -- (8.6,2.6);
\draw[black] (7,8.2) -- (8.6,4.2);
\draw[black] (7,8.2) -- (8.6,5.8);
\draw[black] (2.2,2.2) rectangle (3.8,3);
\draw[black] (2.2,3.8) rectangle (3.8,4.6);
\draw[black] (2.2,5.4) rectangle (3.8,6.2);
\draw[black] (2.2,7.8) rectangle (3.8,8.6);
\path (3,7.2) node[text=black,anchor=base] {\large{}$\vdots$};
\draw[black] (8.6,2.2) rectangle (10.2,3);
\draw[black] (8.6,3.8) rectangle (10.2,4.6);
\draw[black] (8.6,5.4) rectangle (10.2,6.2);
\draw[black] (8.6,7.8) rectangle (10.2,8.6);
\path (9.4,7.2) node[text=black,anchor=base] {\large{}$\vdots$};
\draw[black] (13.4,2.2) rectangle (15,3);
\draw[black] (13.4,3.8) rectangle (15,4.6);
\draw[black] (13.4,5.4) rectangle (15,6.2);
\draw[black] (13.4,7.8) rectangle (15,8.6);
\path (14.2,7.2) node[text=black,anchor=base] {\large{}$\vdots$};
\draw[black] (5.4,2.2) rectangle (7,3);
\draw[black] (5.4,3.8) rectangle (7,4.6);
\draw[black] (5.4,5.4) rectangle (7,6.2);
\draw[black] (5.4,7.8) rectangle (7,8.6);
\path (6.2,7.2) node[text=black,anchor=base] {\large{}$\vdots$};
\path (3,9.2) node[text=black,anchor=base] {\large{}$f$};
\path (4.6,9.2) node[text=black,anchor=base] {\large{}$g$};
\path (3,1.9) node[text=black,anchor=base] {\large{}$t=1$};
\path (6.2,1.9) node[text=black,anchor=base] {\large{}$t=2$};
\path (9.4,1.9) node[text=black,anchor=base] {\large{}$t=3$};
\path (14.2,1.9) node[text=black,anchor=base] {\large{}$t=T$};
\path (1.4,2.7) node[text=black,anchor=base] {\large{}$j_1=1$};
\path (1.4,5.9) node[text=black,anchor=base] {\large{}$j_1=3$};
\path (1.4,4.3) node[text=black,anchor=base] {\large{}$j_1=2$};
\path (1.4,8.3) node[text=black,anchor=base] {\large{}$j_1=J^t$};
\path (3,2.7) node[text=black,anchor=base] {\large{}$b^1_1$};
\path (3,4.3) node[text=black,anchor=base] {\large{}$b^1_2$};
\path (3,5.9) node[text=black,anchor=base] {\large{}$b^1_3$};
\path (6.2,2.7) node[text=black,anchor=base] {\large{}$b^2_1$};
\path (6.2,4.3) node[text=black,anchor=base] {\large{}$b^2_2$};
\path (6.2,5.9) node[text=black,anchor=base] {\large{}$b^2_3$};
\path (9.4,2.7) node[text=black,anchor=base] {\large{}$b^3_1$};
\path (9.4,4.3) node[text=black,anchor=base] {\large{}$b^3_2$};
\path (9.4,5.9) node[text=black,anchor=base] {\large{}$b^3_3$};
\path (14.2,2.7) node[text=black,anchor=base] {\large{}$b^T_1$};
\path (14.2,4.3) node[text=black,anchor=base] {\large{}$b^T_2$};
\path (14.2,5.9) node[text=black,anchor=base] {\large{}$b^T_3$};
\path (3,8.3) node[text=black,anchor=base] {\large{}$b^1_{J^1}$};
\path (6.2,8.3) node[text=black,anchor=base] {\large{}$b^2_{J^2}$};
\path (9.4,8.3) node[text=black,anchor=base] {\large{}$b^3_{J^3}$};
\path (14.2,8.3) node[text=black,anchor=base] {\large{}$b^T_{J^T}$};
\path (11.8,2.7) node[text=black,anchor=base] {\large{}$\cdots$};
\path (11.8,4.3) node[text=black,anchor=base] {\large{}$\cdots$};
\path (11.8,5.9) node[text=black,anchor=base] {\large{}$\cdots$};
\path (11.8,8.3) node[text=black,anchor=base] {\large{}$\cdots$};

\end{tikzpicture}%
      }&
      \raisebox{25pt}{{\large $\times\cdots\times$}}&
      \scalebox{0.2}{
        \begin{tikzpicture}[y=-1cm]

\draw[black] (3.8,4.2) -- (5.4,4.2);
\draw[black] (3.8,2.6) -- (5.4,2.6);
\draw[black] (3.8,5.8) -- (5.4,5.8);
\draw[black] (3.8,8.2) -- (5.4,8.2);
\draw[black] (3.8,2.6) -- (5.4,4.2);
\draw[black] (3.8,2.6) -- (5.4,5.8);
\draw[black] (3.8,2.6) -- (5.4,8.2);
\draw[black] (3.8,4.2) -- (5.4,2.6);
\draw[black] (3.8,4.2) -- (5.4,5.8);
\draw[black] (3.8,4.2) -- (5.4,8.2);
\draw[black] (3.8,5.8) -- (5.4,2.6);
\draw[black] (3.8,5.8) -- (5.4,4.2);
\draw[black] (3.8,5.8) -- (5.4,8.2);
\draw[black] (3.8,8.2) -- (5.4,2.6);
\draw[black] (3.8,8.2) -- (5.4,4.2);
\draw[black] (3.8,8.2) -- (5.4,5.8);
\draw[black] (7,4.2) -- (8.6,4.2);
\draw[black] (7,2.6) -- (8.6,2.6);
\draw[black] (7,5.8) -- (8.6,5.8);
\draw[black] (7,8.2) -- (8.6,8.2);
\draw[black] (7,2.6) -- (8.6,4.2);
\draw[black] (7,2.6) -- (8.6,5.8);
\draw[black] (7,2.6) -- (8.6,8.2);
\draw[black] (7,4.2) -- (8.6,2.6);
\draw[black] (7,4.2) -- (8.6,5.8);
\draw[black] (7,4.2) -- (8.6,8.2);
\draw[black] (7,5.8) -- (8.6,2.6);
\draw[black] (7,5.8) -- (8.6,4.2);
\draw[black] (7,5.8) -- (8.6,8.2);
\draw[black] (7,8.2) -- (8.6,2.6);
\draw[black] (7,8.2) -- (8.6,4.2);
\draw[black] (7,8.2) -- (8.6,5.8);
\draw[black] (2.2,2.2) rectangle (3.8,3);
\draw[black] (2.2,3.8) rectangle (3.8,4.6);
\draw[black] (2.2,5.4) rectangle (3.8,6.2);
\draw[black] (2.2,7.8) rectangle (3.8,8.6);
\path (3,7.2) node[text=black,anchor=base] {\large{}$\vdots$};
\draw[black] (8.6,2.2) rectangle (10.2,3);
\draw[black] (8.6,3.8) rectangle (10.2,4.6);
\draw[black] (8.6,5.4) rectangle (10.2,6.2);
\draw[black] (8.6,7.8) rectangle (10.2,8.6);
\path (9.4,7.2) node[text=black,anchor=base] {\large{}$\vdots$};
\draw[black] (13.4,2.2) rectangle (15,3);
\draw[black] (13.4,3.8) rectangle (15,4.6);
\draw[black] (13.4,5.4) rectangle (15,6.2);
\draw[black] (13.4,7.8) rectangle (15,8.6);
\path (14.2,7.2) node[text=black,anchor=base] {\large{}$\vdots$};
\draw[black] (5.4,2.2) rectangle (7,3);
\draw[black] (5.4,3.8) rectangle (7,4.6);
\draw[black] (5.4,5.4) rectangle (7,6.2);
\draw[black] (5.4,7.8) rectangle (7,8.6);
\path (6.2,7.2) node[text=black,anchor=base] {\large{}$\vdots$};
\path (3,9.2) node[text=black,anchor=base] {\large{}$f$};
\path (4.6,9.2) node[text=black,anchor=base] {\large{}$g$};
\path (3,1.9) node[text=black,anchor=base] {\large{}$t=1$};
\path (6.2,1.9) node[text=black,anchor=base] {\large{}$t=2$};
\path (9.4,1.9) node[text=black,anchor=base] {\large{}$t=3$};
\path (14.2,1.9) node[text=black,anchor=base] {\large{}$t=T$};
\path (1.4,2.7) node[text=black,anchor=base] {\large{}$j_L=1$};
\path (1.4,5.9) node[text=black,anchor=base] {\large{}$j_L=3$};
\path (1.4,4.3) node[text=black,anchor=base] {\large{}$j_L=2$};
\path (1.4,8.3) node[text=black,anchor=base] {\large{}$j_L=J^t$};
\path (3,2.7) node[text=black,anchor=base] {\large{}$b^1_1$};
\path (3,4.3) node[text=black,anchor=base] {\large{}$b^1_2$};
\path (3,5.9) node[text=black,anchor=base] {\large{}$b^1_3$};
\path (6.2,2.7) node[text=black,anchor=base] {\large{}$b^2_1$};
\path (6.2,4.3) node[text=black,anchor=base] {\large{}$b^2_2$};
\path (6.2,5.9) node[text=black,anchor=base] {\large{}$b^2_3$};
\path (9.4,2.7) node[text=black,anchor=base] {\large{}$b^3_1$};
\path (9.4,4.3) node[text=black,anchor=base] {\large{}$b^3_2$};
\path (9.4,5.9) node[text=black,anchor=base] {\large{}$b^3_3$};
\path (14.2,2.7) node[text=black,anchor=base] {\large{}$b^T_1$};
\path (14.2,4.3) node[text=black,anchor=base] {\large{}$b^T_2$};
\path (14.2,5.9) node[text=black,anchor=base] {\large{}$b^T_3$};
\path (3,8.3) node[text=black,anchor=base] {\large{}$b^1_{J^1}$};
\path (6.2,8.3) node[text=black,anchor=base] {\large{}$b^2_{J^2}$};
\path (9.4,8.3) node[text=black,anchor=base] {\large{}$b^3_{J^3}$};
\path (14.2,8.3) node[text=black,anchor=base] {\large{}$b^T_{J^T}$};
\path (11.8,2.7) node[text=black,anchor=base] {\large{}$\cdots$};
\path (11.8,4.3) node[text=black,anchor=base] {\large{}$\cdots$};
\path (11.8,5.9) node[text=black,anchor=base] {\large{}$\cdots$};
\path (11.8,8.3) node[text=black,anchor=base] {\large{}$\cdots$};

\end{tikzpicture}%
      }\\
      &\raisebox{5pt}{{\Large $\times$}}&\\
      \scalebox{0.2}{
        \begin{tikzpicture}[y=-1cm]

\draw[black] (3.8,4.2) -- (5.4,4.2);
\draw[black] (3.8,2.6) -- (5.4,2.6);
\draw[black] (3.8,5.8) -- (5.4,5.8);
\draw[black] (3.8,8.2) -- (5.4,8.2);
\draw[black] (3.8,2.6) -- (5.4,4.2);
\draw[black] (3.8,2.6) -- (5.4,5.8);
\draw[black] (3.8,2.6) -- (5.4,8.2);
\draw[black] (3.8,4.2) -- (5.4,2.6);
\draw[black] (3.8,4.2) -- (5.4,5.8);
\draw[black] (3.8,4.2) -- (5.4,8.2);
\draw[black] (3.8,5.8) -- (5.4,2.6);
\draw[black] (3.8,5.8) -- (5.4,4.2);
\draw[black] (3.8,5.8) -- (5.4,8.2);
\draw[black] (3.8,8.2) -- (5.4,2.6);
\draw[black] (3.8,8.2) -- (5.4,4.2);
\draw[black] (3.8,8.2) -- (5.4,5.8);
\draw[black] (7,4.2) -- (8.6,4.2);
\draw[black] (7,2.6) -- (8.6,2.6);
\draw[black] (7,5.8) -- (8.6,5.8);
\draw[black] (7,8.2) -- (8.6,8.2);
\draw[black] (7,2.6) -- (8.6,4.2);
\draw[black] (7,2.6) -- (8.6,5.8);
\draw[black] (7,2.6) -- (8.6,8.2);
\draw[black] (7,4.2) -- (8.6,2.6);
\draw[black] (7,4.2) -- (8.6,5.8);
\draw[black] (7,4.2) -- (8.6,8.2);
\draw[black] (7,5.8) -- (8.6,2.6);
\draw[black] (7,5.8) -- (8.6,4.2);
\draw[black] (7,5.8) -- (8.6,8.2);
\draw[black] (7,8.2) -- (8.6,2.6);
\draw[black] (7,8.2) -- (8.6,4.2);
\draw[black] (7,8.2) -- (8.6,5.8);
\draw[black] (5.4,2.2) rectangle (7,3);
\draw[black] (5.4,3.8) rectangle (7,4.6);
\draw[black] (5.4,5.4) rectangle (7,6.2);
\draw[black] (5.4,7.8) rectangle (7,8.6);
\path (6.2,7.2) node[text=black,anchor=base] {\large{}$\vdots$};
\draw[black] (8.6,2.2) rectangle (10.2,3);
\draw[black] (8.6,3.8) rectangle (10.2,4.6);
\draw[black] (8.6,5.4) rectangle (10.2,6.2);
\draw[black] (8.6,7.8) rectangle (10.2,8.6);
\path (9.4,7.2) node[text=black,anchor=base] {\large{}$\vdots$};
\draw[black] (2.2,2.2) rectangle (3.8,3);
\draw[black] (2.2,3.8) rectangle (3.8,4.6);
\draw[black] (2.2,5.4) rectangle (3.8,6.2);
\draw[black] (2.2,7.8) rectangle (3.8,8.6);
\path (3,7.2) node[text=black,anchor=base] {\large{}$\vdots$};
\draw[black] (13.4,2.2) rectangle (15,3);
\draw[black] (13.4,3.8) rectangle (15,4.6);
\draw[black] (13.4,5.4) rectangle (15,6.2);
\draw[black] (13.4,7.8) rectangle (15,8.6);
\path (14.2,7.2) node[text=black,anchor=base] {\large{}$\vdots$};
\path (3,1.9) node[text=black,anchor=base] {\large{}$t=1$};
\path (6.2,1.9) node[text=black,anchor=base] {\large{}$t=2$};
\path (9.4,1.9) node[text=black,anchor=base] {\large{}$t=3$};
\path (14.2,1.9) node[text=black,anchor=base] {\large{}$t=T$};
\path (1.4,2.7) node[text=black,anchor=base] {$k_1=1$};
\path (1.4,4.3) node[text=black,anchor=base] {$k_1=2$};
\path (1.4,8.3) node[text=black,anchor=base] {$k_1=K_{s_1}$};
\path (1.4,5.9) node[text=black,anchor=base] {$k_1=3$};
\path (3,9.2) node[text=black,anchor=base] {\large{}$h_{s_1}$};
\path (4.6,9.2) node[text=black,anchor=base] {\large{}$a_{s_1}$};
\path (11.8,2.7) node[text=black,anchor=base] {\large{}$\cdots$};
\path (11.8,4.3) node[text=black,anchor=base] {\large{}$\cdots$};
\path (11.8,5.9) node[text=black,anchor=base] {\large{}$\cdots$};
\path (11.8,8.3) node[text=black,anchor=base] {\large{}$\cdots$};
\path (3,2.7) node[text=black,anchor=base] {\large{}\scalebox{0.45}{$1,b^1_{j^1_{\theta^1_1}},\ldots,b^1_{j^1_{\theta^{I_{s_1}}_1}}$}};
\path (3,4.3) node[text=black,anchor=base] {\large{}\scalebox{0.45}{$2,b^1_{j^1_{\theta^1_1}},\ldots,b^1_{j^1_{\theta^{I_{s_1}}_1}}$}};
\path (3,5.9) node[text=black,anchor=base] {\large{}\scalebox{0.45}{$3,b^1_{j^1_{\theta^1_1}},\ldots,b^1_{j^1_{\theta^{I_{s_1}}_1}}$}};
\path (3,8.3) node[text=black,anchor=base] {\large{}\scalebox{0.45}{$K_{s_1},b^1_{j^1_{\theta^1_1}},\ldots,b^1_{j^1_{\theta^{I_{s_1}}_1}}$}};
\path (6.2,2.7) node[text=black,anchor=base] {\large{}\scalebox{0.45}{$1,b^2_{j^2_{\theta^1_1}},\ldots,b^2_{j^2_{\theta^{I_{s_1}}_1}}$}};
\path (6.2,4.3) node[text=black,anchor=base] {\large{}\scalebox{0.45}{$2,b^2_{j^2_{\theta^1_1}},\ldots,b^2_{j^2_{\theta^{I_{s_1}}_1}}$}};
\path (6.2,5.9) node[text=black,anchor=base] {\large{}\scalebox{0.45}{$3,b^2_{j^2_{\theta^1_1}},\ldots,b^2_{j^2_{\theta^{I_{s_1}}_1}}$}};
\path (6.2,8.3) node[text=black,anchor=base] {\large{}\scalebox{0.45}{$K_{s_1},b^2_{j^2_{\theta^1_1}},\ldots,b^2_{j^2_{\theta^{I_{s_1}}_1}}$}};
\path (9.4,2.7) node[text=black,anchor=base] {\large{}\scalebox{0.45}{$1,b^3_{j^3_{\theta^1_1}},\ldots,b^3_{j^3_{\theta^{I_{s_1}}_1}}$}};
\path (9.4,4.3) node[text=black,anchor=base] {\large{}\scalebox{0.45}{$2,b^3_{j^3_{\theta^1_1}},\ldots,b^3_{j^3_{\theta^{I_{s_1}}_1}}$}};
\path (9.4,5.9) node[text=black,anchor=base] {\large{}\scalebox{0.45}{$3,b^3_{j^3_{\theta^1_1}},\ldots,b^3_{j^3_{\theta^{I_{s_1}}_1}}$}};
\path (9.4,8.3) node[text=black,anchor=base] {\large{}\scalebox{0.45}{$K_{s_1},b^3_{j^3_{\theta^1_1}},\ldots,b^3_{j^3_{\theta^{I_{s_1}}_1}}$}};
\path (14.2,2.7) node[text=black,anchor=base] {\large{}\scalebox{0.45}{$1,b^T_{j^T_{\theta^1_1}},\ldots,b^T_{j^T_{\theta^{I_{s_1}}_1}}$}};
\path (14.2,4.3) node[text=black,anchor=base] {\large{}\scalebox{0.45}{$2,b^T_{j^T_{\theta^1_1}},\ldots,b^T_{j^T_{\theta^{I_{s_1}}_1}}$}};
\path (14.2,5.9) node[text=black,anchor=base] {\large{}\scalebox{0.45}{$3,b^T_{j^T_{\theta^1_1}},\ldots,b^T_{j^T_{\theta^{I_{s_1}}_1}}$}};
\path (14.2,8.3) node[text=black,anchor=base] {\large{}\scalebox{0.45}{$K_{s_1},b^T_{j^T_{\theta^1_1}},\ldots,b^T_{j^T_{\theta^{I_{s_1}}_1}}$}};

\end{tikzpicture}%
      }&
      \raisebox{25pt}{{\large $\times\cdots\times$}}&
      \scalebox{0.2}{
        \begin{tikzpicture}[y=-1cm]

\draw[black] (3.8,4.2) -- (5.4,4.2);
\draw[black] (3.8,2.6) -- (5.4,2.6);
\draw[black] (3.8,5.8) -- (5.4,5.8);
\draw[black] (3.8,8.2) -- (5.4,8.2);
\draw[black] (3.8,2.6) -- (5.4,4.2);
\draw[black] (3.8,2.6) -- (5.4,5.8);
\draw[black] (3.8,2.6) -- (5.4,8.2);
\draw[black] (3.8,4.2) -- (5.4,2.6);
\draw[black] (3.8,4.2) -- (5.4,5.8);
\draw[black] (3.8,4.2) -- (5.4,8.2);
\draw[black] (3.8,5.8) -- (5.4,2.6);
\draw[black] (3.8,5.8) -- (5.4,4.2);
\draw[black] (3.8,5.8) -- (5.4,8.2);
\draw[black] (3.8,8.2) -- (5.4,2.6);
\draw[black] (3.8,8.2) -- (5.4,4.2);
\draw[black] (3.8,8.2) -- (5.4,5.8);
\draw[black] (7,4.2) -- (8.6,4.2);
\draw[black] (7,2.6) -- (8.6,2.6);
\draw[black] (7,5.8) -- (8.6,5.8);
\draw[black] (7,8.2) -- (8.6,8.2);
\draw[black] (7,2.6) -- (8.6,4.2);
\draw[black] (7,2.6) -- (8.6,5.8);
\draw[black] (7,2.6) -- (8.6,8.2);
\draw[black] (7,4.2) -- (8.6,2.6);
\draw[black] (7,4.2) -- (8.6,5.8);
\draw[black] (7,4.2) -- (8.6,8.2);
\draw[black] (7,5.8) -- (8.6,2.6);
\draw[black] (7,5.8) -- (8.6,4.2);
\draw[black] (7,5.8) -- (8.6,8.2);
\draw[black] (7,8.2) -- (8.6,2.6);
\draw[black] (7,8.2) -- (8.6,4.2);
\draw[black] (7,8.2) -- (8.6,5.8);
\draw[black] (5.4,2.2) rectangle (7,3);
\draw[black] (5.4,3.8) rectangle (7,4.6);
\draw[black] (5.4,5.4) rectangle (7,6.2);
\draw[black] (5.4,7.8) rectangle (7,8.6);
\path (6.2,7.2) node[text=black,anchor=base] {\large{}$\vdots$};
\draw[black] (8.6,2.2) rectangle (10.2,3);
\draw[black] (8.6,3.8) rectangle (10.2,4.6);
\draw[black] (8.6,5.4) rectangle (10.2,6.2);
\draw[black] (8.6,7.8) rectangle (10.2,8.6);
\path (9.4,7.2) node[text=black,anchor=base] {\large{}$\vdots$};
\draw[black] (2.2,2.2) rectangle (3.8,3);
\draw[black] (2.2,3.8) rectangle (3.8,4.6);
\draw[black] (2.2,5.4) rectangle (3.8,6.2);
\draw[black] (2.2,7.8) rectangle (3.8,8.6);
\path (3,7.2) node[text=black,anchor=base] {\large{}$\vdots$};
\draw[black] (13.4,2.2) rectangle (15,3);
\draw[black] (13.4,3.8) rectangle (15,4.6);
\draw[black] (13.4,5.4) rectangle (15,6.2);
\draw[black] (13.4,7.8) rectangle (15,8.6);
\path (14.2,7.2) node[text=black,anchor=base] {\large{}$\vdots$};
\path (3,1.9) node[text=black,anchor=base] {\large{}$t=1$};
\path (6.2,1.9) node[text=black,anchor=base] {\large{}$t=2$};
\path (9.4,1.9) node[text=black,anchor=base] {\large{}$t=3$};
\path (14.2,1.9) node[text=black,anchor=base] {\large{}$t=T$};
\path (1.4,2.7) node[text=black,anchor=base] {$k_W=1$};
\path (1.4,4.3) node[text=black,anchor=base] {$k_W=2$};
\path (1.4,8.3) node[text=black,anchor=base] {$k_W=K_{s_W}$};
\path (1.4,5.9) node[text=black,anchor=base] {$k_W=3$};
\path (3,9.2) node[text=black,anchor=base] {\large{}$h_{s_W}$};
\path (4.6,9.2) node[text=black,anchor=base] {\large{}$a_{s_W}$};
\path (11.8,2.7) node[text=black,anchor=base] {\large{}$\cdots$};
\path (11.8,4.3) node[text=black,anchor=base] {\large{}$\cdots$};
\path (11.8,5.9) node[text=black,anchor=base] {\large{}$\cdots$};
\path (11.8,8.3) node[text=black,anchor=base] {\large{}$\cdots$};
\path (3,2.7) node[text=black,anchor=base] {\large{}\scalebox{0.42}{$1,b^1_{j^1_{\theta^1_W}},\ldots,b^1_{j^1_{\theta^{I_{s_W}}_W}}$}};
\path (3,4.3) node[text=black,anchor=base] {\large{}\scalebox{0.42}{$2,b^1_{j^1_{\theta^1_W}},\ldots,b^1_{j^1_{\theta^{I_{s_W}}_W}}$}};
\path (3,5.9) node[text=black,anchor=base] {\large{}\scalebox{0.42}{$3,b^1_{j^1_{\theta^1_W}},\ldots,b^1_{j^1_{\theta^{I_{s_W}}_W}}$}};
\path (3,8.3) node[text=black,anchor=base] {\large{}\scalebox{0.42}{$K_{s_W},b^1_{j^1_{\theta^1_W}},\ldots,b^1_{j^1_{\theta^{I_{s_W}}_W}}$}};
\path (6.2,2.7) node[text=black,anchor=base] {\large{}\scalebox{0.42}{$1,b^2_{j^2_{\theta^1_W}},\ldots,b^2_{j^2_{\theta^{I_{s_W}}_W}}$}};
\path (6.2,4.3) node[text=black,anchor=base] {\large{}\scalebox{0.42}{$2,b^2_{j^2_{\theta^1_W}},\ldots,b^2_{j^2_{\theta^{I_{s_W}}_W}}$}};
\path (6.2,5.9) node[text=black,anchor=base] {\large{}\scalebox{0.42}{$3,b^2_{j^2_{\theta^1_W}},\ldots,b^2_{j^2_{\theta^{I_{s_W}}_W}}$}};
\path (6.2,8.3) node[text=black,anchor=base] {\large{}\scalebox{0.42}{$K_{s_W},b^2_{j^2_{\theta^1_W}},\ldots,b^2_{j^2_{\theta^{I_{s_W}}_W}}$}};
\path (9.4,2.7) node[text=black,anchor=base] {\large{}\scalebox{0.42}{$1,b^3_{j^3_{\theta^1_W}},\ldots,b^3_{j^3_{\theta^{I_{s_W}}_W}}$}};
\path (9.4,4.3) node[text=black,anchor=base] {\large{}\scalebox{0.42}{$2,b^3_{j^3_{\theta^1_W}},\ldots,b^3_{j^3_{\theta^{I_{s_W}}_W}}$}};
\path (9.4,5.9) node[text=black,anchor=base] {\large{}\scalebox{0.42}{$3,b^3_{j^3_{\theta^1_W}},\ldots,b^3_{j^3_{\theta^{I_{s_W}}_W}}$}};
\path (9.4,8.3) node[text=black,anchor=base] {\large{}\scalebox{0.42}{$K_{s_W},b^3_{j^3_{\theta^1_W}},\ldots,b^3_{j^3_{\theta^{I_{s_W}}_W}}$}};
\path (14.2,2.7) node[text=black,anchor=base] {\large{}\scalebox{0.42}{$1,b^T_{j^T_{\theta^1_W}},\ldots,b^T_{j^T_{\theta^{I_{s_W}}_W}}$}};
\path (14.2,4.3) node[text=black,anchor=base] {\large{}\scalebox{0.42}{$2,b^T_{j^T_{\theta^1_W}},\ldots,b^T_{j^T_{\theta^{I_{s_W}}_W}}$}};
\path (14.2,5.9) node[text=black,anchor=base] {\large{}\scalebox{0.42}{$3,b^T_{j^T_{\theta^1_W}},\ldots,b^T_{j^T_{\theta^{I_{s_W}}_W}}$}};
\path (14.2,8.3) node[text=black,anchor=base] {\large{}\scalebox{0.42}{$K_{s_W},b^T_{j^T_{\theta^1_W}},\ldots,b^T_{j^T_{\theta^{I_{s_W}}_W}}$}};

\end{tikzpicture}%
      }\\
      word $1$&&word $W$
    \end{tabular}
  \end{center}
  \vspace*{-2ex}
  \caption{The cross-product lattice used by the sentence tracker, consisting
    of~$L$ tracking lattices and~$W$ event-model lattices.}
  \label{fig:sentence-tracker}
\vspace*{-3ex}
\end{figure}
One would expect that encoding the semantics of a complex sentence such as
\emph{The person to the right of the chair quickly carried the red object
  towards the trash can}, which involves nouns, adjectives, verbs, adverbs, and
spatial-relation and motion prepositions, would provide substantially more
mutual constraint on the \emph{collection} of tracks for the participants than a
single intransitive verb would constrain a single track.
We thus extend the approach described above by incorporating a complex
multi-argument predicate that represents the semantics of an entire sentence
instead of one that only represents the semantics of a single intransitive verb.
This involves formulating the semantics of other parts of speech, in addition
to intransitive verbs, also as HMMs.
We then construct a large cross-product lattice, illustrated in
Fig.~\ref{fig:sentence-tracker}, to support~$L$ tracks and~$W$ words.
Each node in this cross-product lattice represents $L$~detections and the
states for~$W$ words.
To support~$L$ tracks, we subindex each detection index~$j$ as $j_l$ for
track~$l$.
Similarly, to support~$W$ words, we subindex each state index~$k$ as~$k_w$
for word~$w$, the number of states~$K$ for the lexical entry~$s_w$ at word~$w$
as~$K_{s_w}$ and the HMM parameters~$h$ and~$a$ for the lexical entry~$s_w$ at
word~$w$ as~$h_{s_w}$ and~$a_{s_w}$.
The argument-to-track mapping~$\theta^i_w$ specifies the track
that fills argument~$i$ of word~$w$, where~$I_{s_w}$ specifies the arity, the
number of arguments, of the lexical entry~$s_w$ at word~$w$.
We then seek a path through this cross-product lattice that optimizes
\vspace*{-2ex}
\begin{equation*}
  \max_{\substack{j^1_1,\ldots,j^T_1\\j^1_L,\ldots,j^T_L\\
      k^1_1,\ldots,k^T_1\\k^1_W,\ldots,k^T_W}}\hspace*{-1ex}
  \begin{array}[t]{l}
    \displaystyle
    \sum_{l=1}^L\;
    \left(\sum_{t=1}^Tf(b^t_{j^t_l})+
    \sum_{t=2}^Tg(b^{t-1}_{j^{t-1}_l},b^t_{j^t_l})\right)\\[-0.2ex]
    {}+\hspace*{-1ex}
    \begin{array}[t]{l}
      \displaystyle
      \sum_{w=1}^W
      \left(\sum_{t=1}^Th_{s_w}(k^t_w,b^t_{j^t_{\theta^1_w}},\ldots,b^t_{j^t_{\theta^{I_{s_w}}_w}})\right.\\[-1.5ex]
      \displaystyle
      \left.\;\;\;\;\;\;\;\;\;{}+\sum_{t=2}^Ta_{s_w}(k^{t-1}_w,k^t_w)\right)
    \end{array}
  \end{array}
\vspace*{-1.5ex}
\end{equation*}
This can also be computed in polynomial time using the Viterbi algorithm.
This describes a method by which the function
$\mathcal{S}:(\mathbf{B},\mathbf{s},\Lambda)\mapsto(\tau,\mathbf{J})$,
discussed earlier, can be computed, where $\mathbf{B}$~is the collection of
detections~$b^t_j$ and~$\mathbf{J}$ is the collection of detection
indices~$j^t_l$.

The complexity of the sentence tracker is $O(T(J^LK^W)^2)$ in time and
$O(J^LK^W)$ in space, where~$T$ is the number of frames in the video, $W$~is
the number of words in the sentence~$\mathbf{s}$, $L$~is the number of
participants, $J=\max\left\{ J^1,\ldots,J^T\right\}$, where $J^t$ is the number
of detections considered in frame~$t$,
and~$K=\max\left\{K_{s_1},\ldots,K_{s_W}\right\}$.
In practice, $J\leq5$, $L\leq4$, and $K=1$ for all but verbs and motion
prepositions of which there are typically no more than three.
With such, the method takes less than a second.

\vspace*{-1.5ex}
\section{Natural-Language Semantics}
\label{sec:semantics}
\vspace*{-1ex}

The sentence tracker uniformly represents the semantics of words in all parts
of speech, namely nouns, adjectives, verbs, adverbs, and prepositions (both
those that describe spatial relations and those that describe motion), as HMMs.
Finite-state recognizers (FSMs) are a special case of HMMs where the transition
matrices~$a$ and the output models~$h$ are 0/1, which become $-\infty$/0 in
$\log$ space.
Here, we formulate the semantics of a small fragment of English consisting
of~17 lexical items (5~nouns, 2~adjectives, 4~verbs, 2~adverbs,
2~spatial-relation prepositions, and 2~motion prepositions), by hand, as FSMs.
We do so to focus on what one can do with this approach as discussed in
section~\ref{sec:experiments}.
It is particularly enlightening that the FSMs we use are perspicuous and
clearly encode pretheoretic human intuitions about word semantics.
But nothing turns on the use of hand-coded FSMs.
Our framework, as described above, supports HMMs.

\begin{table*}[t]
  \centering\ra{1.1}
  \begin{tabular}{@{}c@{\hspace*{10ex}}c@{}}
    \begin{tabular}{@{}cc@{}}
      (a)&
      \scalebox{0.60}{\begin{tabular}{@{}r@{$\;\rightarrow\;$}l@{}}
        S & NP VP\\
        NP & D [A] N [PP]\\
        D & \emph{an} \textbar\ \emph{the}\\
        A & \emph{blue} \textbar\ \emph{red}\\
        N & \emph{person} \textbar\ \emph{backpack} \textbar\ \emph{chair} \textbar\ \emph{trash can} \textbar\ \emph{object}\\
        PP & P NP\\
        P & \emph{to the left of} \textbar\ \emph{to the right of}\\
        VP & V NP [Adv] [$\text{PP}_{\text{M}}$]\\
        V & \emph{approached} \textbar\ \emph{carried} \textbar\ \emph{picked up} \textbar\ \emph{put down}\\
        Adv & \emph{quickly} \textbar\ \emph{slowly}\\
        $\text{PP}_{\text{M}}$ & $\text{P}_{\text{M}}$ NP\\
        $\text{P}_{\text{M}}$ & \emph{towards} \textbar\ \emph{away from}
      \end{tabular}}\\[11ex]
      (b)&
      \scalebox{0.60}{\begin{tabular}{@{}r@{$:\;$}l@{}}
          \emph{to the left of} &
          $\{\textrm{agent},\textrm{patient},\textrm{source},\textrm{goal},\textrm{referent}\},\{\textrm{referent}\}$\\
          \emph{to the right of} &
          $\{\textrm{agent},\textrm{patient},\textrm{source},\textrm{goal},\textrm{referent}\},\{\textrm{referent}\}$\\
          \emph{approached} & $\{\textrm{agent}\},\{\textrm{goal}\}$\\
          \emph{carried} & $\{\textrm{agent}\},\{\textrm{patient}\}$\\
          \emph{picked up} & $\{\textrm{agent}\},\{\textrm{patient}\}$\\
          \emph{put down} & $\{\textrm{agent}\},\{\textrm{patient}\}$\\
          \emph{towards} & $\{\textrm{agent},\textrm{patient}\},\{\textrm{goal}\}$\\
          \emph{away from} &
          $\{\textrm{agent},\textrm{patient}\},\{\textrm{source}\}$\\
          other &
          $\{\textrm{agent},\textrm{patient},\textrm{source},\textrm{goal},\textrm{referent}\}$\\
        \end{tabular}}
    \end{tabular}&
  \begin{tabular}{@{}cc@{}}
    (c)&
    \scalebox{0.63}{\begin{tabular}{@{}r@{\,}l@{.\hspace{1ex}}l@{}}
        1&{\red a}&\emph{The {\red backpack} approached the trash can.}\\
        &{\green b}&\emph{The {\green chair} approached the trash can.}\\
        2&{\red a}&\emph{The {\red red} object approached the trash can.}\\
        &{\green b}&\emph{The {\green blue} object approached the trash can.}\\
        3&{\red a}&\emph{The person to the {\red left} of the trash can put down an object.}\\
        &{\green b}&\emph{The person to the {\green right} of the trash can put down an object.}\\
        4&{\red a}&\emph{The person put down the {\red trash can}.}\\
        &{\green b}&\emph{The person put down the {\green backpack}.}\\
        5&{\red a}&\emph{The person carried the {\red red} object.}\\
        &{\green b}&\emph{The person carried the {\green blue} object.}\\
        6&{\red a}&\emph{The person picked up an object to the {\red left} of the trash can.}\\
        &{\green b}&\emph{The person picked up an object to the {\green right} of the trash can.}\\
        7&{\red a}&\emph{The person {\red picked up} an object.}\\
        &{\green b}&\emph{The person {\green put down} an object.}\\
        8&{\red a}&\emph{The person picked up an object {\red quickly}.}\\
        &{\green b}&\emph{The person picked up an object {\green slowly}.}\\
        9&{\red a}&\emph{The person carried an object {\red towards} the trash can.}\\
        &{\green b}&\emph{The person carried an object {\green away from} the trash can.}\\
        1&0&\emph{The backpack approached the chair.}\\
        1&1&\emph{The red object approached the chair.}\\
        1&2&\emph{The person put down the chair.}
      \end{tabular}}
    \end{tabular}
  \end{tabular}
  \vspace*{0.2ex}
  \caption{(a) The grammar for our lexicon of 17 lexical entries (5~nouns,
    2~adjectives, 4~verbs, 2~adverbs, 2~spatial-relation prepositions, and
    2~motion prepositions).
    Note that the grammar allows for infinite recursion.
    (b) Specification of the number of arguments for each
    word and the roles such arguments refer to.
    (c) A selection of sentences drawn from the grammar based on which we
    collected our corpus.}
  \label{tab:language}
\vspace*{-4ex}
\end{table*}

Nouns (\eg\ \emph{person}) may be represented by constructing static FSMs over
discrete features, such as detector class.
Adjectives (\eg\ \emph{red}, \emph{tall}, and \emph{big}) may be represented as
static FSMs that describe select properties of the detections for a single
participant, such as color, shape, or size, independent of other features of
the overall event.
Intransitive verbs (\eg\ \emph{bounce}) may be represented as FSMs that
describe the changing motion characteristics of a single participant, such as
\emph{moving downward} followed by \emph{moving upward}.
Transitive verbs (\eg\ \emph{approach}) may be represented as FSMs that
describe the changing relative motion characteristics of two participants, such
as \emph{moving closer}.
Adverbs (\eg\ \emph{slowly} and \emph{quickly}) may be represented by FSMs that
describe the velocity of a single participant, independent of the direction of
motion.
Spatial-relation prepositions (\eg\ \emph{to the left of}) may be represented
as static FSMs that describe the relative position of two participants.
Motion prepositions (\eg\ \emph{towards} and \emph{away from}) may be
represented as FSMs that describe the changing relative position of two
participants.
As is often the case, even simple static properties, such as detector class,
object color, shape, and size, spatial relations, and direction of motion,
might hold only for a portion of an event.
We handle such temporal uncertainty by incorporating garbage states into the
FSMs that always accept and do not affect the scores computed.
This also allows for alignment between multiple words in a temporal interval
during a longer aggregate event.
We formulate the FSMs for specifying the word meanings as regular expressions
over predicates computed from detections.
The particular set of regular expressions and associated predicates that are
used in the experiments are given in Table~\ref{tab:predicates}.
The predicates are formulated around a number of primitive functions.
The function $\textit{avgFlow}(b)$ computes a vector that represents the
average optical flow inside the detection~$b$.
The functions $x(b)$, $\textit{model}(b)$, and $\textit{hue}(b)$ return the
$x$-coordinate of the center of~$b$, its object class, and the average hue of
the pixels inside~$b$ respectively.
The function $\textit{fwdProj}(b)$ displaces~$b$ by the average optical flow
inside~$b$.
The functions $\angle$ and $\textit{angleSep}$ determine the angular component
of a given vector and angular distance between two angular arguments
respectively.
The function $\textit{normal}$ computes a normal unit vector for a given
vector.
The argument~$v$ to $\textsc{noJitter}$ denotes a specified direction
represented as a 2D unit vector in that direction.
Regular expressions are formed around predicates as atoms.
A given regular expression must be formed solely from output models of the same
arity and denotes an FSM, \ie\ an HMM with a 0/1 transition matrix and output
model, which become $-\infty$/0 in $\log$ space.
We use $R^{\{n,\}}\define R \stackrel{n}{\cdots} R\;R^{*}$ to indicate that~$R$
must be repeated at least~$n$ times and
$R^{[n,]}\define(\textsc{R}\;[\textsc{true}])^{\{n,\}}$ to indicate that~$R$
must be repeated at least~$n$ times but can optionally have a single frame of
noise between each repetition.
This allows for some flexibility in the models.

\begin{table*}
  \centering
  \scalebox{0.62}{\begin{tabular}{@{}c@{}}
      \begin{tabular}{@{\hspace{0ex}}c@{\hspace{-1ex}}c@{\hspace{-1ex}}c@{\hspace{0ex}}}
        \midrule Constants & Simple Predicates & Complex Predicates\\[-0.5ex] \midrule \addlinespace[1ex]
        \begin{math}
          \begin{array}[t]{r@{\;\define\;}l}
            \textsc{xBoundary} & 300\textsc{px}\\
            \textsc{nextTo} & 50\textsc{px}\\
            \Delta\textsc{static} & 6\textsc{px}\\
            \Delta\textsc{jump} & 30\textsc{px}\\
            \Delta\textsc{quick} & 80\textsc{px}\\
            \Delta\textsc{slow} & 30\textsc{px}\\
            \Delta\textsc{closing} & 10\textsc{px}\\
            \Delta\textsc{direction} & 30\degr\\
            \Delta\textsc{hue} & 30\degr
          \end{array}
        \end{math} &
        \begin{math}
          \begin{array}[t]{r@{\;\define\;}l}
            \textsc{noJitter}(b,v) &
            \lVert\textit{avgFlow}(b)\cdot v\rVert\le\Delta\textsc{jump}\\
            \textsc{alike}(b_1,b_2) &
            \textit{model}(b_1)=\textit{model}(b_2)\\
            \textsc{close}(b_1,b_2) &
            \lvert x(b_1)-x(b_2)\rvert<\textsc{xBoundary}\\
            \textsc{far}(b_1,b_2) &
            \lvert x(b_1)-x(b_2)\rvert\ge\textsc{xBoundary}\\
            \textsc{left}(b_1,b_2) &
            0<x(b_2)-x(b_1)\le\textsc{nextTo}\\
            \textsc{right}(b_1,b_2) &
            0<x(b_1)-x(b_2)\le\textsc{nextTo}\\
            \textsc{hasColor}(b,\textrm{hue}) &
            \textit{angleSep}(\textit{hue}(b),\textrm{hue})\le
            \Delta\textsc{hue}\\
            \textsc{stationary}(b) &
            \lVert\textit{avgFlow}(b)\rVert\le\Delta\textsc{static}\\
            \textsc{quick}(b) &
            \lVert\textit{avgFlow}(b)\rVert\ge\Delta\textsc{quick}\\
            \textsc{slow}(b) &
            \lVert\textit{avgFlow}(b)\rVert\le\Delta\textsc{slow}\\
            \textsc{person}(b) & \textit{model}(b)=\textbf{person}\\
            \textsc{backpack}(b) & \textit{model}(b)=\textbf{backpack}\\
            \textsc{chair}(b) & \textit{model}(b)=\textbf{chair}\\
            \textsc{trashcan}(b) & \textit{model}(b)=\textbf{trashcan}\\
            \textsc{blue}(b) & \textsc{hasColor}(b,225^\circ)\\
            \textsc{red}(b) & \textsc{hasColor}(b,0^\circ)
          \end{array}
        \end{math}&
        \begin{math}
          \begin{array}[t]{r@{\;\define\;}l}
            \textsc{stationaryClose}(b_1,b_2) &
            \textsc{stationary}(b_1)\conjunct
            \textsc{stationary}(b_2)\conjunct
            \neg\textsc{alike}(b_1,b_2)\conjunct
            \textsc{close}(b_1,b_2)\\
            \textsc{stationaryFar}(b_1,b_2) &
            \textsc{stationary}(b_1)\conjunct
            \textsc{stationary}(b_2)\conjunct
            \neg\textsc{alike}(b_1,b_2)\conjunct
            \textsc{far}(b_1,b_2)\\
            \textsc{closer}(b_1,b_2) &
            \lvert x(b_1)-x(b_2)\rvert>
            \lvert x(\textit{fwdProj}(b_1))-x(b_2)\rvert+
            \Delta\textsc{closing}\\
            \textsc{farther}(b_1,b_2) &
            \lvert x(b_1)-x(b_2)\rvert<
            \lvert x(\textit{fwdProj}(b_1))-x(b_2)\rvert+
            \Delta\textsc{closing}\\
            \textsc{moveCloser}(b_1,b_2) &
            \textsc{noJitter}(b_1,(0,1))\conjunct
            \textsc{noJitter}(b_2,(0,1))\conjunct
            \textsc{closer}(b_1,b_2)\\
            \textsc{moveFarther}(b_1,b_2) &
            \textsc{noJitter}(b_1,(0,1))\conjunct
            \textsc{noJitter}(b_2,(0,1))\conjunct
            \textsc{farther}(b_1,b_2)\\
            \textsc{inAngle}(b,v)&
            \textit{angleSep}(\angle\textit{avgFlow}(b),\angle v)<
            \Delta\textsc{angle}\\
            \textsc{inDirection}(b,v) &
            \textsc{noJitter}(b,\normal(v))\conjunct
            \neg\textsc{stationary}(b)\conjunct
            \textsc{inAngle}(b,v)\\
            \textsc{approaching}(b_1,b_2) &
            \neg\textsc{alike}(b_1,b_2)\conjunct
            \textsc{stationary}(b_2)\conjunct
            \textsc{moveCloser}(b_1,b_2)\\
            \textsc{carry}(b_1,b_2,v) &
            \textsc{person}(b_1)\conjunct
            \neg\textsc{alike}(b_1,b_2)\conjunct
            \textsc{inDirection}(b_1,v)\conjunct
            \textsc{inDirection}(b_2,v)\\
            \textsc{carrying}(b_1,b_2) &
            \textsc{carry}(b_1,b_2,(0,1))\disjunct
            \textsc{carry}(b_1,b_2,(0,-1))\\
            \textsc{departing}(b_1,b_2) &
            \neg\textsc{alike}(b_1,b_2)\conjunct
            \textsc{stationary}(b_2)\conjunct
            \textsc{moveFarther}(b_1,b_2)\\
            \textsc{pickingUp}(b_1,b_2) &
            \textsc{person}(b_1)\conjunct
            \neg\textsc{alike}(b_1,b_2)\conjunct
            \textsc{stationary}(b_1)\conjunct
            \textsc{inDirection}(b_2,(0,1))\\
            \textsc{puttingDown}(b_1,b_2) &
            \textsc{person}(b_1)\conjunct
            \neg\textsc{alike}(b_1,b_2)\conjunct
            \textsc{stationary}(b_1)\conjunct
            \textsc{inDirection}(b_2,(0,-1))\\
          \end{array}
        \end{math}
      \end{tabular}\\
      \addlinespace[0.5ex] \midrule Regular Expressions\\ \midrule \addlinespace[1ex]
      \begin{tabular}{@{}c@{\hspace{3pt}}c@{\hspace{3pt}}c@{}}
        \begin{math}
          \begin{array}[t]{r@{\;\define\;}l}
            \lambda_{\textit{person}} & \textsc{person}\plus\\
            \lambda_{\textit{backpack}} & \textsc{backpack}\plus\\
            \lambda_{\textit{chair}} & \textsc{chair}\plus\\
            \lambda_{\textit{trash can}} & \textsc{trashcan} \plus\\
            \lambda_{\textit{object}} &
            (\textsc{backpack}\;|\;\textsc{chair}\;|\;\textsc{trashcan})
            \plus
          \end{array}
        \end{math}&
        \begin{math}
          \begin{array}[t]{r@{\;\define\;}l}
            \lambda_{\textit{blue}} & \textsc{blue}\plus\\
            \lambda_{\textit{red}} & \textsc{red}\plus\\
            \lambda_{\textit{quickly}} &
            \textsc{true}\plus\;\textsc{quick}\dagr\;\textsc{true}\plus\\
            \lambda_{\textit{slowly}} &
            \textsc{true}\plus\;\textsc{slow}\dagr\;\textsc{true}\plus\\
            \lambda_{\textit{to the left of}} & \textsc{left}\plus\\
            \lambda_{\textit{to the right of}} & \textsc{right}\plus
          \end{array}
        \end{math}&
        \begin{math}
          \begin{array}[t]{r@{\;\define\;}l}
            \lambda_{\textit{approached}} &
            \textsc{stationaryFar}\plus\;
            \textsc{approaching}\dagr\;
            \textsc{stationaryClose}\plus\\
            \lambda_{\textit{carried}} &
            \textsc{stationaryClose}\plus\;
            \textsc{carrying}\dagr\;
            \textsc{stationaryClose}\plus\\
            \lambda_{\textit{picked up}} &
            \textsc{stationaryClose}\plus\;
            \textsc{pickingUp}\dagr\;
            \textsc{stationaryClose}\plus\\
            \lambda_{\textit{put down}} &
            \textsc{stationaryClose}\plus\;
            \textsc{puttingDown}\dagr\;
            \textsc{stationaryClose}\plus\\
            \lambda_{\textit{towards}} &
            \textsc{stationaryFar}\plus\;
            \textsc{approaching}\dagr\;
            \textsc{stationaryClose}\plus\\
            \lambda_{\textit{away from}} &
            \textsc{stationaryClose}\plus\;
            \textsc{departing}\dagr\;
            \textsc{stationaryFar}\plus\\[1ex]
          \end{array}
        \end{math}
      \end{tabular}
    \end{tabular}}
  \caption{The finite-state recognizers corresponding to the lexicon in
    Table~\protect\ref{tab:language}(a).}
  \label{tab:predicates}
\vspace*{-4ex}
\end{table*}

A sentence may describe an activity involving multiple tracks, where different
(collections of) tracks fill the arguments of different words.
This gives rise to the requirement of compositional semantics: dealing with the
mappings from arguments to tracks.
Argument-to-track assignment is a function $\Theta:\mathbf{s}\mapsto(L,\theta)$
that maps a sentence~$\mathbf{s}$ to the number~$L$ of participants and the
argument-to-track mapping~$\theta^i_w$.
The mapping specifies which tracks fill which arguments of which words in the
sentence and is mediated by a grammar and a specification of the argument arity
and role types for the words in the lexicon.
Given a sentence, say \emph{The person to the right of the chair picked up the
  backpack}, along with the grammar specified in Table~\ref{tab:language}(a)
and the lexicon specified in Tables~\ref{tab:language}(b)
and~\ref{tab:predicates}, it would yield a mapping corresponding to the
following formula.
\vspace*{-1.3ex}
\begin{equation*}
  \begin{array}[t]{l}
    \textsc{person}(P) \wedge
    \textsc{toTheRightOf}(P,Q)\wedge
    \textsc{chair}(Q)\\
    \wedge\ \textsc{pickedUp}(P,R)\wedge
    \textsc{backpack}(R)
  \end{array}
\vspace*{-1.3ex}
\end{equation*}

\noindent
To do so, we first construct a parse tree of the sentence~$\mathbf{s}$ given the
grammar, using a recursive-descent parser.
For each word, we then determine from the parse tree, which words in the
sentence are determined to be its \textsl{dependents} in the sense of
\textsl{government}, and how many such \textsl{dependents} exist, from the
lexicon specified in Table~\ref{tab:language}(b).
For example, the dependents of \emph{to the right of} are determined to be
\emph{person} and \emph{chair}, filling its first and second arguments
respectively.
Moreover, we determine a consistent assignment of roles, one of agent,
patient, source, goal, and referent, for each participant track that fills the
word arguments, from the allowed roles specified for that word
and argument in the lexicon.
Here, $P$, $Q$, and~$R$ are participants that play the agent, referent, and
patient roles respectively.

\vspace*{-2ex}
\section{Experimental Evaluation}
\label{sec:experiments}
\vspace*{-1ex}

The sentence tracker supports three distinct capabilities.
It can take sentences as input and focus the attention of a tracker, it can
take video as input and produce sentential descriptions as output, and it can
perform content-based video retrieval given a sentential input query.
To evaluate the first three, we filmed a corpus of~94 short video clips, of
varying length, in~3 different outdoor environments.
The camera was moved for each video clip so that the varying background
precluded unanticipated confounds.
These video clips, filmed with a variety of actors, each depicted one or more of
the~21 sentences from Table~\ref{tab:language}(c).
The depiction, from video clip to video clip, varied in scene layout and the
actor(s) performing the event.
The corpus was carefully constructed in a number of ways.
First, many video clips depict more than one sentence.
In particular, many video clips depict simultaneous distinct events.
Second, each sentence is depicted by multiple video clips.
Third the corpus was constructed with minimal pairs: pairs of video clips whose
depicted sentences differ in exactly one word.
These minimal pairs are indicated as the `a' and `b' variants of sentences 1--9
in Table~\ref{tab:language}(c).
That varying word was carefully chosen to span all parts of speech and all
sentential positions: sentence~1 varies subject noun, sentence~2 varies subject
adjective, sentence~3 varies subject preposition, sentence~4 varies object
noun, sentence~5 varies object adjective, sentence~6 varies object preposition,
sentence~7 varies verb, sentence~8 varies adverb, and sentence~9 varies
motion preposition.
We filmed our own corpus as we are unaware of any existing corpora that exhibit
the above properties.
We annotated each of the~94 clips with ground truth judgments for each of
the~21 sentences, indicating whether the given clip depicted the given
sentence.
This set of~1974 judgments was used for the following analyses.

\vspace*{-1ex}
\subsection{Focus of Attention}
\label{subsec:foa}
\vspace*{-1ex}
Tracking is traditionally performed using cues from motion, object detection,
or manual initialization on an object of interest.
However, in the case of a cluttered scene involving multiple activities
occurring simultaneously, there can be many moving objects, many instances of
the same object class, and perhaps even multiple simultaneously occurring
instances of the same event class.
This presents a significant obstacle to the efficacy of existing methods in
such scenarios.
To alleviate this problem, one can decide which objects to track based on which
ones participate in a target event.

\begin{figure*}
  \centering
  \begin{tabular}
    {@{}c@{\hspace{6pt}}c@{\hspace{6pt}}c@{\hspace{6pt}}c@{}}
    \includegraphics[width=0.18\textwidth]{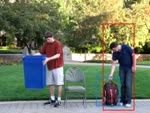}&
    \includegraphics[width=0.18\textwidth]{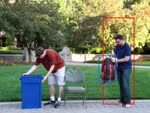}&
    \includegraphics[width=0.18\textwidth]{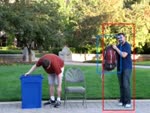}&
    \includegraphics[width=0.18\textwidth]{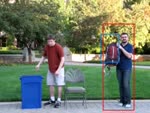}\\
    \multicolumn{4}{@{}l}{\emph{The person picked up an object.}}\\
    \includegraphics[width=0.18\textwidth]{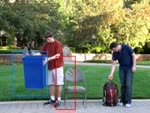}&
    \includegraphics[width=0.18\textwidth]{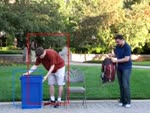}&
    \includegraphics[width=0.18\textwidth]{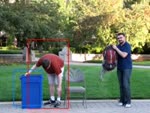}&
    \includegraphics[width=0.18\textwidth]{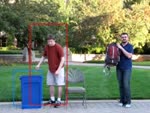}\\
    \multicolumn{4}{@{}l}{\emph{The person put down an object.}}\\[0.3ex]
  \end{tabular}
  \caption{Sentence-guided focus of attention: different sets of tracks for
    the same video clip produced under guidance of different sentences.
    Here, and in Figs.~\ref{fig:generation} and~\ref{fig:retrieval},
    the red box denotes the agent, the blue box denotes the patient, the violet
    box denotes the source, the turquoise box denotes the goal, and the green
    box denotes the referent.
    These roles are determined automatically.}
  \label{fig:inference}
  \vspace*{-2ex}
\end{figure*}

The sentence tracker can focus its attention on just those objects that
participate in an event specified by a sentential description.
Such a description can differentiate between different simultaneous events
taking place between many moving objects in the scene using descriptions
constructed out of a variety of parts of speech: nouns to specify object class,
adjectives to specify object properties, verbs to specify events, adverbs to
specify motion properties, and prepositions to specify (changing) spatial
relations between objects.
Furthermore, such a sentential description can even differentiate which objects
to track based on the role that they play in an event: agent, patient, source,
goal, or referent.
Fig.~\ref{fig:inference} demonstrates this ability: different tracks are
produced for the same video clip that depicts multiple simultaneous events when
focused with different sentences.

We further evaluated this ability on all~9 minimal pairs, collectively applied
to all~24 suitable video clips in our corpus.
For 21~of these, both sentences in the minimal pair yielded tracks deemed to be
correct depictions.
Our
website\footnote{\scriptsize\url{http://aql.ecn.purdue.edu/~qobi/cccp/cvpr2014.html}}
includes example video clips for all~9 minimal pairs.

\vspace*{-1ex}
\subsection{Generation}
\label{subsec:generation}
\vspace*{-1ex}
Much of the prior work on generating sentences to describe images
\citep{Jie2009, Farhadi2010, Li2011a, Yang2011, Gupta2012, Mitchell2012} and
video \citep{Kojima2002, Tena2007, Barbu2012a, Khan2012, Wang2012} uses
special-purpose natural-language-generation methods.
We can instead use the ability of the sentence tracker to score a sentence
paired with a video clip as a general-purpose natural-language generator by
searching for the highest-scoring sentence for a given video clip.
However, this has a problem.
Scores decrease with longer word sequences and greater numbers of tracks that
result from such.
This is because both~$f$ and~$g$ are mapped to $\log$ space, \ie\ $(-\infty,0]$,
via sigmoids, to match~$h$ and~$a$, which are $\log$ probabilities.
So we don't actually search for the highest-scoring sentence, which would bias
the process towards short sentences.
Instead, we seek complex sentences that are true of the video clip as they are
more informative.

Nominally, this search process would be intractable since the space of possible
sentences can be huge and even infinite.
However, we can use beam search to get an approximate answer.
This is possible because the sentence tracker can score any word sequence, not
just complete phrases or sentences.
We can select the top-scoring single-word sequences and then repeatedly
extend the top-scoring $W$-word sequences, by one word, to select the
top-scoring $W+1$-word sequences, subject to the constraint that these
$W+1$-word sequences are grammatical sentences or can be extended to
grammatical sentences by insertion of additional words.
We terminate the search process when the \defoccur{contraction threshold}, the
ratio between the score of a sequence and the score of the sequence expanding
from it, drops below a specified value and the sequence being expanded is a
complete sentence.
This contraction threshold controls complexity of the generated sentence.

When restricted to FSMs, $h$~and~$a$ will be 0/1, which become $-\infty$/0 in
$\log$ space.
Thus increase in the number of words can only decrease a score to $-\infty$,
meaning that a sequence of words no-longer describes a video clip.
Since we seek sentences that do, we terminate the above beam-search process
before the score goes to $-\infty$.
In this case, there is no approximation: a beam search maintaining all $W$-word
sequences with finite score yields the highest-scoring sentence before the
contraction threshold is met.

\begin{figure*}
  \centering
  \begin{tabular}
    {@{}c@{\hspace{6pt}}c@{\hspace{6pt}}c@{\hspace{6pt}}c@{}}
    \includegraphics[width=0.18\textwidth]{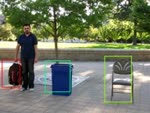}&
    \includegraphics[width=0.18\textwidth]{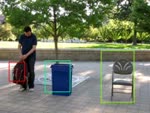}&
    \includegraphics[width=0.18\textwidth]{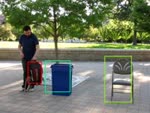}&
    \includegraphics[width=0.18\textwidth]{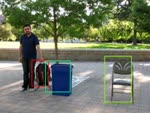}\\
    \multicolumn{4}{@{}l}{\emph{The backpack to the left of the chair approached the trash can.}}\\
    \includegraphics[width=0.18\textwidth]{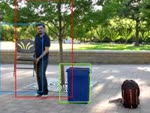}&
    \includegraphics[width=0.18\textwidth]{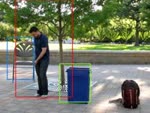}&
    \includegraphics[width=0.18\textwidth]{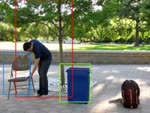}&
    \includegraphics[width=0.18\textwidth]{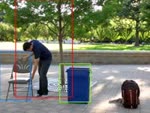}\\
    \multicolumn{4}{@{}l}{\emph{The person to the left of the trash can put down
        the chair.}}\\[0.3ex]
  \end{tabular}
  \caption{Generation of sentential description: constructing the best-scoring
    sentence for each video clip through a beam search.}
  \label{fig:generation}
\vspace*{-4ex}
\end{figure*}

\setcounter{footnote}{0}

To evaluate this approach, we searched the space of sentences generated by the
grammar in Table~\ref{tab:language}(a) to find the top-scoring sentence for
each of the~94 video clips in our corpus.
Note that the grammar generates an infinite number of sentences due to
recursion in NP.\@
Even restricting the grammar to eliminate NP recursion yields a space of
147,123,874,800 sentences.
Despite not restricting the grammar in this fashion, we are able to effectively
find good descriptions of the video clips.
We evaluated the accuracy of the sentence tracker in generating descriptions
for our entire corpus, for multiple contraction thresholds.
Accuracy was computed as the percentage of the~94 clips for which generated
descriptions were deemed to describe the video by human judges.
Contraction thresholds of 0.95, 0.90, and 0.85 yielded accuracies of
67.02\%, 71.27\%, and 64.89\% respectively.
We demonstrate examples of this approach in Fig.~\ref{fig:generation}.
Our website\footnotemark\ contains additional examples.

\setcounter{footnote}{0}

\vspace*{-1ex}
\subsection{Retrieval}
\label{subsec:retrieval}
\begin{figure*}
  \centering
  \begin{tabular}
    {@{}c@{\hspace{6pt}}c@{\hspace{6pt}}c@{\hspace{6pt}}c@{}}
    \includegraphics[width=0.18\textwidth]{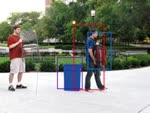}&
    \includegraphics[width=0.18\textwidth]{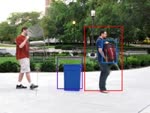}&
    \includegraphics[width=0.18\textwidth]{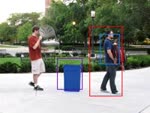}&
    \includegraphics[width=0.18\textwidth]{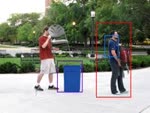}\\
    \multicolumn{4}{@{}l}{\emph{The person carried an object away from the trash
      can.}}\\
    \includegraphics[width=0.18\textwidth]{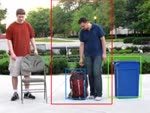}&
    \includegraphics[width=0.18\textwidth]{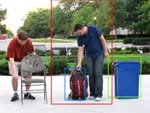}&
    \includegraphics[width=0.18\textwidth]{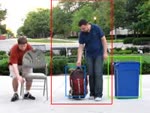}&
    \includegraphics[width=0.18\textwidth]{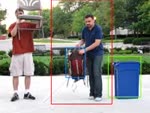}\\
    \multicolumn{4}{@{}l}{\emph{The person picked up an object to the left of
        the trash can.}}\\[0.3ex]
  \end{tabular}
  \caption{Sentential-query-based video search: returning the best-scoring
    video clip, in a corpus of~94 video clips, for a given sentence.}
  \label{fig:retrieval}
\vspace*{-4ex}
\end{figure*}

\vspace*{-1ex}
The availability of vast video corpora, such as on YouTube, has created a
rapidly growing demand for content-based video search and retrieval.
The existing systems, however, only provide a means to search via human-provided
captions.
The inefficacy of such an approach is evident.
Attempting to search for even simple queries such as \emph{pick up} or
\emph{put down} yields surprisingly poor results, let alone searching for more
complex queries such as \emph{person approached horse}.
Furthermore, some prior work on content-based video-retrieval systems, like
Sivic and Zisserman \cite{Sivic2003}, search only for objects and other prior
work, like Laptev \etal\ \cite{laptev:08}, search only for events.
Even combining such to support conjunctive queries for video clips with
specified collections of objects jointly with a specified event, would not
effectively rule out video clips where the specified objects did not play a
role in the event or played different roles in the event.
For example, it could not rule out a video clip depicting a person jumping next
to a stationary ball for a query \emph{ball bounce} or distinguish between the
queries \emph{person approached horse} and \emph{horse approached person}.
The sentence tracker exhibits the ability to serve as the basis of a much
better video search and retrieval tool, one that performs content-based search
with complex sentential queries to find precise semantically relevant clips,
as demonstrated in Fig.~\ref{fig:retrieval}.
Our website\footnotemark\ contains the top three scoring video clips for each
query sentence from Table~\ref{tab:language}(c).

\setcounter{footnote}{0}

To evaluate this approach, we scored every video clip in our corpus against
every sentence in Table~\ref{tab:language}(c), rank ordering the video clips
for each sentence, yielding the following statistics over the~1974 scores.
\vspace*{-1ex}
\begin{center}\small\sl
  \begin{tabular}{@{}p{0.37\textwidth}@{\hspace{2ex}}r@{}}
    chance that a random clip depicts a given sentence & 13.12\%\\
    top-scoring clip depicts the given sentence & 94.68\%\\
    $\ge$~1 of top 3 clips depicts the given sentence &
    100.00\%\\
  \end{tabular}
\vspace*{-1ex}
\end{center}
Our website\footnotemark\ contains all~94 video clips and all~1974 scores.
The judgment of whether a video clip depicted a given sentence was made using
our annotation.
We conducted an additional evaluation with this annotation.
One can threshold the sentence-tracker score to yield a binary predicate on
video-sentence pairs.
We performed 4-fold cross validation on our corpus, selecting the threshold for
each fold that maximized accuracy of this predicate, relative to the
annotation, on 75\% of the video clips and evaluating the accuracy with this
selected threshold on the remaining 25\%.
This yielded an average accuracy of~86.88\%.

\vspace*{-3ex}
\section{Conclusion}
\label{sec:conclusion}
\vspace*{-2ex}
We have presented a novel framework that utilizes the compositional structure
of events and the compositional structure of language to drive a semantically
meaningful and targeted approach towards activity recognition.
This multi-modal framework integrates low-level visual components, such as
object detectors, with high-level semantic information in the form of
sentential descriptions in natural language.
This is facilitated by the shared structure of detection-based tracking, which
incorporates the low-level object-detector components, and of finite-state
recognizers, which incorporate the semantics of the words in a lexicon.

We demonstrated the utility and expressiveness of our framework by performing
three separate tasks on our corpus, requiring no training or annotation, simply
by leveraging our framework in different manners.
The first, sentence-guided focus of attention, showcases the ability to focus
the attention of a tracker on the activity described in a sentence, indicating
the capability to identify such subtle distinctions as between \emph{The person
  picked up the chair to the left of the trash can} and \emph{The person picked
  up the chair to the right of the trash can}.
The second, generation of sentential description of video, showcases the
ability to produce a complex description of a video clip, involving multiple
parts of speech, by performing an efficient search for the best description
through the space of all possible descriptions.
The final task, query-based video search, showcases the ability to perform
content-based video search and retrieval, allowing for such distinctions as
between \emph{The person approached the trash can} and \emph{The trash can
  approached the person}.
\ifcvprfinal
\vspace*{-3ex}
\paragraph{Acknowledgments}
This research was supported, in part, by ARL, under Cooperative Agreement Number
W911NF-10-2-0060, and the Center for Brains, Minds and Machines, funded by NSF
STC award CCF-1231216.
The views and conclusions contained in this document are those of the authors
and do not represent the official policies, either express or implied, of ARL
or the U.S. Government.
The U.S. Government is authorized to reproduce and distribute reprints for
Government purposes, notwithstanding any copyright notation herein.
\fi
\vspace*{-2ex}
{\small
\bibliographystyle{ieee}
\bibliography{arxiv2013e}
}

\end{document}